%% file: arxiv.tex
\documentclass[runningheads]{llncs}
\usepackage[T1]{fontenc}
\usepackage{graphicx}
\usepackage{multirow}
\usepackage{color}
\usepackage{amsmath}
\usepackage{booktabs}
\usepackage{amssymb}
\begin{document}
%
\title{
What Matters in Virtual Try-Off? Dual-UNet Diffusion Models for Garment Reconstruction \thanks{Paper accepted at ICPR 2026.}
}
%
%
\author{Loc-Phat Truong\inst{1,2} \and
Meysam Madadi\inst{1,2} \and
Sergio Escalera\inst{1,2}}
\authorrunning{L.P. Truong et al.}
%

\institute{Computer Vision Center, Spain \and
Universitat de Barcelona, Spain\\
\email{phattruong@ub.edu}\\
\email{mmadadi@ub.edu}\\
\email{sescalera@ub.edu}
}
\titlerunning{What Matters in Virtual Try-Off?}
\maketitle              
\begin{figure}[!h]
    \centering
    \includegraphics[width=\linewidth]{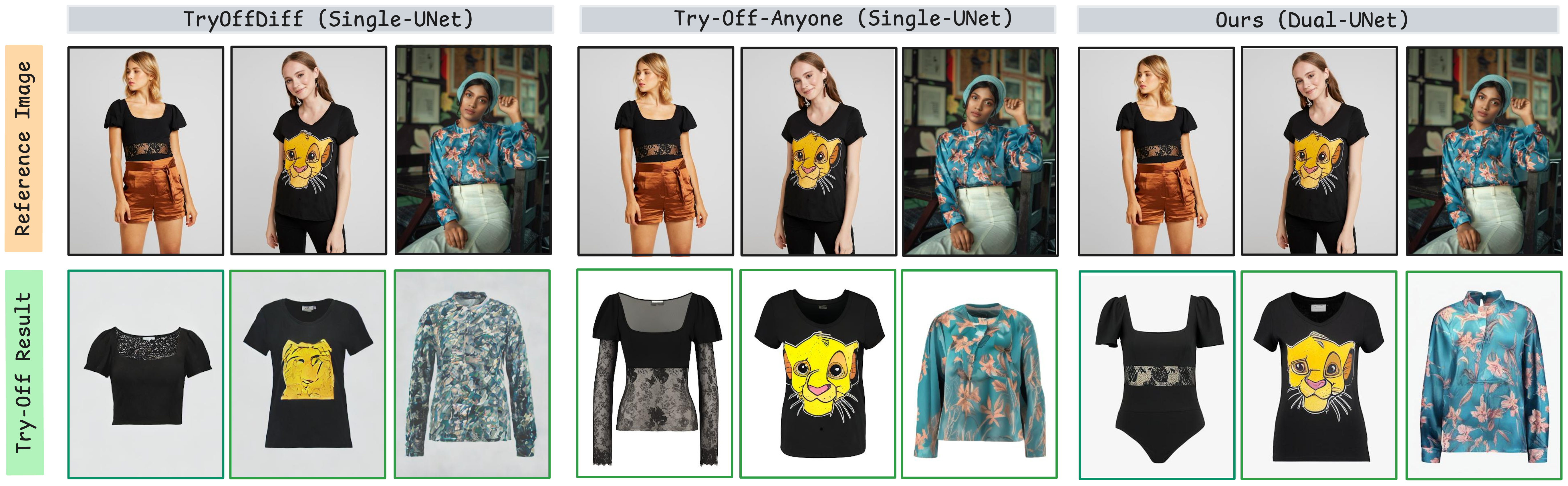}
    \caption{State-of-the-art single-UNet, TryOffDiff~\cite{velioglu_tryoffdiff_2024} and Try-Off-Anyone~\cite{xarchakos_tryoffanyone_2025}, vs. our adapted Dual-UNet try-off results. Our Dual-UNet architecture achieves high-quality and realistic garment generation while precisely preserving fine-grained details across diverse datasets.
    }
    \label{fig:title_im}
\end{figure}

\input{sec/0_abstract}    
\input{sec/1_intro}
\input{sec/2_sota}
\input{sec/3_method}
\input{sec/4_experiment}
\input{sec/5_conclusion}
%
%
%
%
%
%
\bibliographystyle{splncs04}
\bibliography{main}
%





\clearpage
\appendix
\section*{Supplementary Material}
\input{supp/0}

\input{supp/1}
\input{supp/2}
\input{supp/3}
\input{supp/4}
\input{supp/5}
\input{supp/6}
\end{document}

%% file: sec/0_abstract.tex
\begin{abstract}
Virtual Try-On (VTON) has 
seen rapid advancements, providing a strong foundation for generative fashion tasks. 
However, the inverse problem, Virtual Try-Off (VTOFF)-aimed at reconstructing the canonical garment from a draped-on image-remains a less understood domain, distinct from the heavily researched field of VTON.
In this work, we seek to establish a robust architectural foundation for VTOFF by studying and adapting various diffusion-based strategies from VTON and general Latent Diffusion Models (LDMs).
We focus our investigation on the Dual-UNet Diffusion Model architecture 
and analyze three axes of design: (i) Generation Backbone: comparing Stable Diffusion variants; (ii) Conditioning: ablating different mask designs, masked/unmasked inputs for image conditioning, and the utility of high-level semantic features; and (iii) Losses and Training Strategies: evaluating the impact of the auxiliary attention-based loss, perceptual objectives and multi-stage curriculum schedules.
Extensive experiments reveal trade-offs across various configuration options. Evaluated on VITON-HD and DressCode datasets, our framework achieves state-of-the-art performance with a drop of 9.5\% on the primary metric DISTS and competitive performance on LPIPS, FID, KID, and SSIM, providing both stronger baselines and insights to guide future Virtual Try-Off research.
\keywords{Virtual Try-Off  \and Virtual Try-On \and Latent Diffusion Model \and Curriculum Learning.}
\end{abstract}

%% file: sec/1_intro.tex
\section{Introduction}
\label{sec:intro}

Image-based Virtual Try-Off (VTOFF) is the inverse problem of Image-based Virtual Try-On (VTON), which seeks to reconstruct the canonical garment from the worn version on human pose, enabling applications such as product retrieval or large-scale fashion dataset.
VTOFF is also an effective complement to the emerging person-to-person VTON task~\cite{wang2024fldm,wang2025fw}, which aims to transfer garments between individuals, by first reconstructing the target catalogue garment for the use in the typical VTON pipeline.
Not only it can expand the practical usage scenario for users and facilitate the dataset construction for VTON, but the closed loop of VTOFF-VTON also has the potential to improve VTON by enhancing the features shared between both tasks~\cite{wang2024fldm,wang2025fw}.

However, VTOFF is more challenging than VTON due to the partial view and occluded or deformed regions in specific body poses, hindering an accurate generation of the invisible garment appearance and shape.
Despite advances in VTON, VTOFF has started gaining traction recently in a few works~\cite{zeng2020tilegan,velioglu_tryoffdiff_2024,xarchakos_tryoffanyone_2025}. One of the earliest efforts, TileGAN~\cite{zeng2020tilegan}, employed Conditional Generative Adversarial Network (CGAN)~\cite{mirza2014conditional} to generate tiled clothing image from model image. However, it struggles to produce high-quality results due to the lack of powerful foundational model and large-scale datasets. 
More recent approaches have started adopting Latent Diffusion Models (LDMs)~\cite{rombach2022high}, used in VTON, for higher quality output~\cite{velioglu_tryoffdiff_2024,xarchakos_tryoffanyone_2025,shen_igr_2024}.
Single-UNet architectures~\cite{velioglu_tryoffdiff_2024,xarchakos_tryoffanyone_2025} were adopted from CatVTON~\cite{chong_catvton_2025} due to its efficiency without performance sacrifice, while Dual-UNet architectures~\cite{shen_igr_2024} apply IDM-VTON~\cite{choi_improving_2025} to better encode high- and low-level semantic features of garment image.
Despite the superior performance of Single- vs. Dual-UNet architectures in VTON, they show poor results in VTOFF. As one can see in Fig.~\ref{fig:title_im}, the realism and accuracy of cloth shape and texture remain unsatisfied in Single-UNet architectures, suggesting that the adaptation of existing VTON architectural strategies needs a thorough investigation to determine what truly matters for VTOFF's distinct characteristics.

\begin{figure}[!h]
    \centering
    \includegraphics[width=0.8\linewidth]{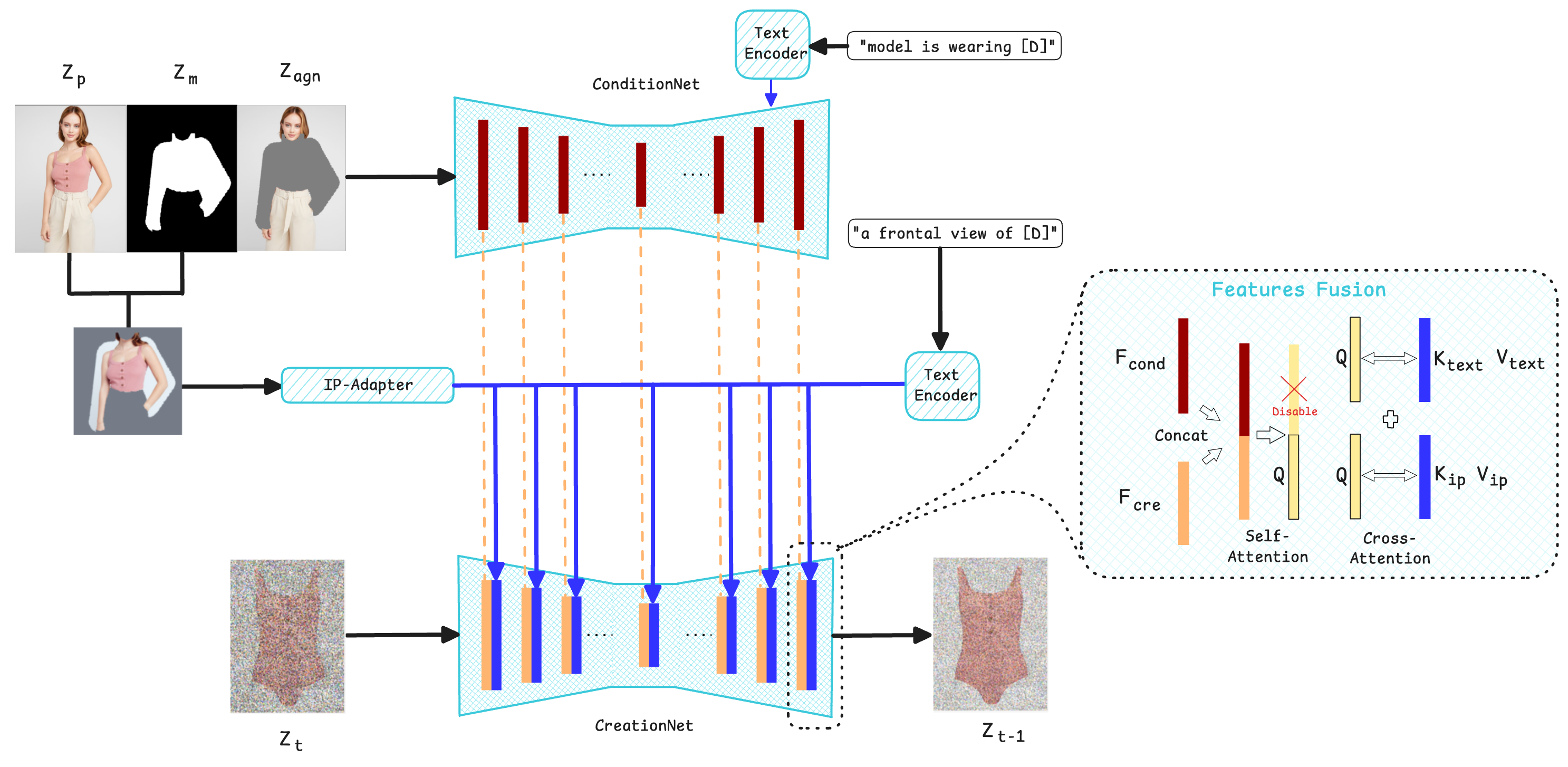}
    \caption{
    \textbf{Overview of the Dual-UNet Diffusion Model framework for VTOFF}. 
    The framework has two branches: 
    (1) Generation branch: CreationNet, a denoising UNet, generates the garment image; 
    (2) Conditioning branch: includes IP-Adapter~\cite{ye2023ip} and a text encoder for high-level features, and ConditionNet for low-level features.
    VTON styled input, the channel concatenation of VAE-encoded latents of the person, mask, and cloth-agnostic image, goes into the ConditionNet, while isolated-garment image is fed into the IP-Adapter.
    High-level (garment description [$D$], IP-Adapter tokens) and low-level (ConditionNet) features fuse via self-attention and cross-attention to condition CreationNet.
    }
    \label{fig:dualunet_arch}
\end{figure}

In this paper, we 
focus on the adaptation of
the Dual-UNet architecture (see Fig. \ref{fig:dualunet_arch}), built upon Stable Diffusion and conditioned by multi-level garment features. 
These consist of high-level representations from image and text prompts, and low-level features from a reference UNet architecturally identical to the main generation UNet.
Second, we evaluate different training objectives. The Leffa loss~\cite{zhou2025learning} is used as an auxiliary training task to regularize the spatial alignment of keys and queries between the intermediate attention layers of the generation and reference UNets. 
This regularization is effective to prevent distortion in fine-grained details such as textures, text, and logos. 
We also analyze the employment of perceptual losses to address unrealistic artifacts.
Third, since both denoising UNets contain a very large number of parameters, training them simultaneously introduces complex interactions and computational overhead, leading to underperforming results.
To address this, we leverage a curriculum learning scheme for interleaved enhancement of both UNets’ capabilities. 
Overall, our approach successfully produces canonical garment reconstructions with precise inference of cloth shape and components, while also improving the alignment of fine-grained details between the input image and the reconstructed result 
as visualized in Fig.~\ref{fig:title_im}
. To summarize, our main contributions are: (1) We present 
a design analysis of Dual-UNet Diffusion Model architectures for Virtual Try-Off
. (2) We systematically explore the utilization of auxiliary modules and training strategies to prevent fine-grained details distortion and enhance all network components without collapsing. (3) We achieve state-of-the-art performance on DISTS, FID, KID, and competitive results on other relevant metrics such as LPIPS, SSIM on the VITON-HD~\cite{choi2021viton} and DressCode~\cite{morelli2022dresscode} dataset.

%% file: sec/2_sota.tex
\section{Related Works}
\textbf{Conditional Latent Diffusion Models}
LDMs~\cite{rombach2022high}, especially the widely-used Stable Diffusion (SD) (e.g.,~SD v1.5~\cite{rombach2022high}, SDXL~\cite{podell2023sdxl}), have become the de facto solution for both Virtual Try-On \cite{morelli2023ladi,gou2023taming,kim_stable_2024,choi_improving_2025,chong_catvton_2025,zhou2025learning} and a few Virtual Try-Off early works \cite{velioglu_tryoffdiff_2024,xarchakos_tryoffanyone_2025,shen_igr_2024}. 
Prior research \cite{zhang2023adding,mou2024t2i,ye2023ip,yang_paint_2022} extends beyond text conditioning by guiding LDMs with reference subjects to enhance consistency and preserve fine-grained details.
ControlNet \cite{zhang2023adding} guides by spatial conditions (e.g.,~human pose) using a copied trainable network.
Similarly, IP-Adapter \cite{ye2023ip} extracts semantic features from references through a pretrained CLIP \cite{radford2021learning} and a small network.
Paint-by-Example \cite{yang_paint_2022} establishes an inpainting pipeline for LDMs, where a masked region is filled by guiding the diffusion model with an image exemplar.

\textbf{Image-based Virtual Try-On}
Existing approaches \cite{jetchev2017conditional,han_viton_2018,wang2018toward,lee2019viton,cui2021dressing,han2019clothflow,zhu2023tryondiffusion,morelli2023ladi,gou2023taming,kim_stable_2024,choi_improving_2025,chong_catvton_2025,zhou2025learning}, have undergone several paradigm shifts including the transition from warping-based methods to warping-free methods, or from Generative Adversarial Network (GAN) \cite{goodfellow2014generative} to Diffusion Model \cite{ho2020denoising}. 
The latest trend approaches VTON as an exemplar-based inpainting problem \cite{yang_paint_2022} employing LDMs to naturally 
synthesize the garment on human body.
StableVITON \cite{kim_stable_2024} follows ControlNet \cite{zhang2023adding} and leverages a copied trainable SD encoder to efficiently utilize local garment features. 
In contrast, IDM-VTON \cite{choi_improving_2025} leverages a frozen UNet for low-level garment features. High-level semantic features are extracted by IP-Adapter \cite{ye2023ip} and text description, while human pose provides additional control signal along with mask.
CatVTON \cite{chong_catvton_2025} spatially concatenates the person image and mask, and efficiently perform self-attention in a single UNet without high-level semantic references.
Leffa \cite{zhou2025learning} mitigates garment identity inaccuracy caused by attention misalignment between the two UNets. It regularizes the main UNet by constructing a flow field from the attention map to warp the reference garment, using a soft-constraint reconstruction loss with the ground truth.
By reviewing VTON architectures and their behavior, we draw insights that inform the construction of our VTOFF framework.

\textbf{Virtual Try-Off}
The idea of disentangling garment representation was first introduced as an auxiliary direction to improve VTON quality \cite{zhang2021vtgun,wang2024fldm}.
However, producing high-quality unwarped cloths is itself a challenging task as it involves both warping and reconstruction tasks. 
Zeng et al. (2020) \cite{zeng2020tilegan} presented the earliest attempt to reconstruct tiled garment from the dressed person image with a two-stage coarse-to-fine GAN, but the method fails to produce high-quality results. 
TryOffDiff \cite{velioglu_tryoffdiff_2024} pioneers the use of higher capacity LDMs for VTOFF and also sets the official task settings with notations, dataset, metrics, and a baseline. 
It follows IP-Adapter \cite{ye2023ip}, extracting garment features through SigLIP \cite{zhai2023sigmoid} followed by a transformer-based \cite{vaswani2017attention} adapter to guide the LDMs.
TryOffAnyone \cite{xarchakos_tryoffanyone_2025} adopts CatVTON \cite{chong_catvton_2025} architecture, concatenating the person image and inpainting mask without using additional conditions.
Both works successfully produce higher quality results and outperform naive adaptation from VTON, but they still suffer distortion in garment structural (e.g.,~in neckline, waist, button) or miss perceptual (e.g.,~color, texture) detail.  
IGR \cite{shen_igr_2024} improves details extraction by combining IP-Adapter \cite{ye2023ip} and the second pretrained diffusion UNet, similar to IDM-VTON \cite{choi_improving_2025}. 
However, these approaches do not thoroughly examine how different inputs, conditions, and training strategies affect VTOFF — an analysis we undertake in this study.


%% file: sec/3_method.tex
\section{\textbf{Method}}

\subsection{\textbf{Dual-UNet Diffusion Model}}
\label{sec:dualunet}

Prior VTON works~\cite{xu_ootdiffusion_2025,choi_improving_2025,zhou2025learning} transform SD into a Dual-UNet Framework in two steps:

First, a new UNet is created: ConditionNet $\theta_{cond}$ (a \textit{base} SD extracting features from the garment $I_c$) which is architecturally identical to CreationNet $\theta_{cre}$ (an \textit{inpainting} SD that denoises a corrupted latent concatenated with the mask $I_m$ and cloth-agnostic person $I_{agn} = I_p\cdot(1-I_m)$, where $I_p$ denotes the person image).
Optionally, IP-adapter \cite{ye2023ip} extract high-level features from the garment image $I_c$ along with textual features from a CLIP \cite{radford2021learning} text encoder.

Second, low-level ConditionNet's features interact with CreationNet through self-attention mechanism, which the augmented features corresponding to CreationNet is then refined by high-level features through cross-attention.

In our model, we adopt the same architecture, consisting of Conditioning and Generation branches, but modify the input accordingly for VTOFF setting as shown in Fig. \ref{fig:dualunet_arch}.
We use the SD \textit{inpainting} pipeline with its input convention for ConditionNet to exploit the capability of garment spatial feature extraction.
Specifically, we concatenate the VAE encoded latents $z_p$, $z_m$, and $z_{agn}$ of the person image $I_p$, binary mask $I_m$, and the cloth-agnostic person image $I_{agn}$.
Next, we obtain image semantic features via IP-Adapter
. To ensure these features focus optimally on the garment region, we crop the person image based on the mask bounding box and then resize it to $224 \times 224$ to match with the CLIP image encoder of IP-Adapter.
In addition, we extract semantic features from the garment description $D$. For CreationNet, $D$ is modified to the prompt “a frontal view of [$D$]”, whereas for ConditionNet, it is “model is wearing [$D$]”. 
Then, CreationNet is incorporated with those features by self-attention and cross-attention mechanism and trained with the garment image $I_c$ as groundtruth.
Based on this framework, we expand the design axes to investigate model behavior and optimize performance, which are organized as follows: Generation Branch (Sec. \ref{sec:generation_branch}), Conditioning Branch (Sec. \ref{sec:conditioning_branch}), and Training (Sec. \ref{sec:lossandcurriculum}).
Moreover, we inherit the description $D$ from IDM-VTON annotation. Considering this data can vary in real-world situations, we conduct a study to test our framework behavior when training with various forms of description $D$. See Supp.~Sec.~5
for more insights.

\subsubsection{Generation Branch Design}
\label{sec:generation_branch}
\textit{Stable Diffusion Base vs Stable Diffusion Inpainting.}
For CreationNet, a \textit{base} SD variant is effective because, unlike VTON, VTOFF does not require preserving surrounding context around a mask.
However, we hypothesize that providing a white background prior can shorten the optimization path due to the background bias in VITON-HD dataset.
To test this hypothesis, we switch CreationNet into the SD $inpainting$ variant receiving the mask prior that combines a mask and a masked central background. 
Specifically, we create a white background $I_{white}$ and a background mask $I_{bgmask}$ that preserves a narrow border (10\% each side) while removing the central area, yielding $I_{maskedbg} = I_{white} \cdot (1 - I_{bgmask})$.

\textit{Stable Diffusion v1.5 vs Stable Diffusion SDXL.}
SD version 1.5 and SDXL are the two most widely used SD models, differing in architecture and generative performance. 
We examine the behavior of both models by maintaining the same dual-branch framework and only switch SD version. 
Note that the version of IP-Adapter and Text Encoder are also modified according to each version. We do not consider higher versions such as SD 3.5 due to fundamental differences: Diffusion Transformer (DiT) \cite{peebles2023scalable} architecture and Flow Matching \cite{lipman2023flow} training.  

\subsubsection{Conditioning Branch Design}
\label{sec:conditioning_branch}
\textit{Truncate IP-Adapter and Text Encoder.}
High-level semantic features are crucial in VTOFF, especially taking into account the high degree of freedom when generating the results. 
To validate the impact and necessity of the high-level features, we compare two settings: (1) the dual-UNet alone, and (2) the full framework with IP-Adapter and text encoder. 

\textit{Fit vs Dilated Binary Mask.}
The mask directly affects generation quality in various VTON pipelines, as it guides $inpainting$ signals on the relevant regions. 
Given 
this utility in VTOFF's conditioning branch, we compare different mask types: the popular dilated mask, which provides contextual information by covering the person's neck and arms along with the cloth, and the fit mask, which focuses exclusively on the garment's segmentation.

\textit{Masked vs Unmasked Cloth-Focused Image for IP-Adapter.}
For IP-Adapter in VTOFF, we follow the same pre-processing protocol as in VTON, but utilize the isolated-garment image $I_{iso} = I_p \cdot I_m$.
However, we investigate whether this isolation is necessary, specifically evaluating the sensitivity of the IP-Adapter's high-level semantic features to information outside the garment and assessing whether surrounding context provides useful cues or introduces distraction.
\subsection{\textbf{Auxiliary Modules \& Training Strategies}}
\label{sec:lossandcurriculum}
\subsubsection{Diffusion Loss for VTOFF}
is defined as follow:
\begin{equation}
\mathcal{L}_{\text{diff}} = 
\mathbb{E}_{z_0, \epsilon \sim \mathcal{N}(0, I), t} 
\Big[ \| \epsilon - \epsilon_\theta(z_t, t, c) \|_2^2 \Big],
\end{equation}
where $z_0$ denotes the ground-truth VAE encoded latent of $I_c$, $z_t$ is the noisy latent at timestep $t$, $\epsilon$ is Gaussian noise, and $\epsilon_\theta(\cdot)$ is the noise prediction from the UNet conditioned on $c$ (e.g.,~text), and timestep $t \sim \text{Uniform}(\{1, \dots, T\})$.
\subsubsection{Learning Flow Fields in Attention Loss (Leffa)}
We observe that VTOFF exhibits Dual-UNet's attention layers misalignment (Supp.~Sec.~2
), leading to fine-grained detail distortion, similar to what Zhou et al. ~\cite{zhou2025learning} showcase in VTON.
Leffa loss is an attention-guided warping constraint that forces the model to learn an accurate spatial mapping between the person image and the generated garment. 
Specifically, the attention map $A_l$ produced by each Dual-UNet's intermediate attention layer is used to generate a flow field. This flow field is then upsampled into pixel-space and used via grid sampling to warp the isolated garment image ($I_{\text{iso}}$).
The loss $\mathcal{L}_{\text{Leffa}}$ is then the $L_2$ distance between this warped image and the ground-truth garment ($I_c$), summed across selected layers $L$:
\begin{equation}
\mathcal{L}_{\text{Leffa}} = \sum_{l=1}^{L} \| \text{grid\_sampling}\Big(I_{\text{iso}}, \, \text{upsample}(A_l \odot C)\Big) - I_c \|_2^2
\end{equation}
where $C$ is a normalized coordinate map.
Leffa is only enabled in the later stages of training to avoid early-stage performance harm, and we adopt it within our curriculum described in Sec. \ref{sec:currilearn}.
For more implementation and hyperparameters details, we refer the reader to the original Leffa work.


\subsubsection{Perceptual Loss}
To address unrealistic details, we experiment with two perceptual objectives, ELatentLPIPS \cite{kang2024diffusion2gan} and LPL \cite{berrada2025boosting}, defined in general as the loss 
\begin{equation}
\mathcal{L}_{perceptual}=\delta\big( \phi(z_{0}),\phi(\hat{z}_0) \big)
\end{equation}
where $\hat{z}_0$ is the estimated original latent from noise prediction $\epsilon_\theta(\cdot)$ during training, $\phi(\cdot)$ is a pretrained perceptual network and $\delta$ is a distance metric.
\subsubsection{Final Combined Loss} 
Our final loss is a combination of diffusion loss and different auxiliary losses with their corresponding balancing terms (e.g.,~$\lambda_{perceptual}$ and ~$\lambda_{Leffa}$):
\begin{equation}
    \mathcal{L}_{\text{final}} = \mathcal{L}_{\text{diff}} + \lambda_{\text{Leffa}} \mathcal{L}_{\text{Leffa}}  + \lambda_{\text{perceptual}} \mathcal{L}_{\text{perceptual}} 
\end{equation}
\subsubsection{Curriculum Learning}
\label{sec:currilearn}
Training the Dual-UNet framework with its complex architecture, conditioning mechanism, and auxiliary losses is nontrivial and prone to instability. 
To address this, we design a curriculum that progressively integrates components into training to enhance the entire network without sub-optimality. 
We denote the trainable components as follows:
(\textit{cre\_ip}): CreationNet + IP-Adapter trainable, ConditionNet frozen; 
(\textit{cond}): CreationNet + IP-Adapter frozen, ConditionNet trainable; 
(\textit{joint}): all three components trainable.
We begin by optimizing \textit{cre\_ip} with $\mathcal{L}_{\text{diff}}$, leveraging ConditionNet's pretrained inpainting prior to stabilize CreationNet training. Note that we disable Leffa loss in this stage to prevent early harms and improve generalization. Then we enable Leffa loss and train \textit{cre\_ip} one more round.
Finally we freeze it and finetune \textit{cond} to boost the accuracy of fine-grained details in this stage.
In summary, our complete training curriculum consists of three stages: 
(Stage 1): \textit{cre\_ip}, w/o Leffa, 130 epochs; 
(Stage 2): \textit{cre\_ip}, with Leffa, 130 epochs; 
(Stage 3): \textit{cond}, with Leffa, 130 epochs.
We describe curriculum variants further in Sec. \ref{sec:exp_curr_design}

%% file: sec/4_experiment.tex
\section{\textbf{Experiments}}
\subsection{\textbf{Dataset, Metrics, and Experimental Setup}}
\textbf{Dataset}
\footnotetext[1]{KID is multiplied by $10^3$ to bring it to a similar scale as other metrics and is the mean value over 10 evaluation runs.}
To evaluate our framework’s capability, we conduct experiments on VITON-HD dataset \cite{choi2021viton} at 512x384 resolution for all exploratory ablations to fairly compare to TryOffDiff \cite{velioglu_tryoffdiff_2024} and Try-Off-Anyone \cite{xarchakos_tryoffanyone_2025}.
Then we use the best configuration to run final training at 1024x768 resolution and on DressCode dataset \cite{morelli2022dresscode} to fairly compare to IGR \cite{shen_igr_2024} and Try-Off-Diff, respectively. 
We further examine the generalization of models trained on VITON-HD with a self-collected out-of-distribution dataset called Unsplash Real-World dataset.
More details about datasets can be found in Supp.~Sec.~1
.

\textbf{Metrics}
\label{sec:exp_metrics}
Follow TryOffDiff \cite{velioglu_tryoffdiff_2024}, we adopt SSIM, LPIPS, FID, KID, and DISTS to comprehensively evaluate the model performance, with DISTS serving as the primary metric since it simultaneously measure structural and textural similarity. 
Note that we calculate DISTS on the full output size, which avoid the reduced quality of the resizing into $341\times256$ established in TryOffDiff. 
In addition, to effectively interpret the metrics and solidify the robustness of DISTS for this task, we study the reflection of common failures in VTOFF on quantitative numbers.
See the implementation and insights in Supp.~Sec.~3
.
\begin{table}[t]
\centering
\caption{Comparison of the Dual-UNet architectural design ablations as presented in Sec.~\ref{sec:dualunet}. Note \textbf{bold} indicates the best value}
\tiny
\begin{tabular}{lccccccc}
\toprule
\textbf{Architecture Choice} & \textbf{SSIM $\uparrow$} & \textbf{LPIPS $\downarrow$} & \textbf{DISTS $\downarrow$} & \textbf{FID $\downarrow$}& \textbf{KID\footnotemark[1]  $\downarrow$} \\
\midrule
\emph{inpainting} & 73.77 & 31.83 & 22.18 & 11.32 & 1.99 \\
\textbf{\emph{base}} & \textbf{73.94} & \textbf{30.14} & \textbf{20.90} & \textbf{10.17} & \textbf{1.89} \\
\midrule
SDv1.5 & 69.16 & 39.89 & 25.06 & 12.55 & 2.66 \\
\textbf{SDXL} & \textbf{73.94} & \textbf{30.14} & \textbf{20.90} & \textbf{10.17} & \textbf{1.89} \\
\midrule
Truncated & 69.70 & 38.11 & 30.60 & 18.40 & 4.09 \\
\textbf{IP-Adapter + Text} & \textbf{73.94} & \textbf{30.14} & \textbf{20.90} & \textbf{10.17} & \textbf{1.89} \\
\bottomrule
\end{tabular}
\label{tab:architecture_abl}
\end{table}

\begin{table}[t]
\centering
\caption{Comparisons related to masking ablations, including the type of mask and IP-Adapter input masking strategies.}
\tiny
\begin{tabular}{lccccccc}
\toprule
\textbf{Masking Operation} & \textbf{SSIM $\uparrow$} & \textbf{LPIPS $\downarrow$} & \textbf{DISTS $\downarrow$} & \textbf{FID $\downarrow$}& \textbf{KID\footnotemark[1]  $\downarrow$} \\
\midrule
Fit mask & \textbf{74.20} & 30.96 & 21.30 & 10.58 & 2.11 \\
\textbf{Dilated mask} & 73.94 & \textbf{30.14} & \textbf{20.90} & \textbf{10.17} & \textbf{1.89} \\
\midrule
Unmasked IP & 73.57 & \textbf{30.00} & 20.96 & 10.31 & 2.12\\
\textbf{Masked IP} & \textbf{73.94} & 30.14 & \textbf{20.90} & \textbf{10.17} & \textbf{1.89} \\
\bottomrule
\end{tabular}
\label{tab:masking_abl}
\end{table} 

\begin{table}[t]
\centering
\caption{Comparison of training curricula. Curr 6, with Color Jitter in Stage 1 only
, achieves the best performance on DISTS and FID.}
\tiny
\begin{tabular}{lccccccc}
\toprule
\textbf{Curriculum} & \textbf{Color Aug?} & \textbf{SSIM $\uparrow$}  & \textbf{LPIPS $\downarrow$} & \textbf{DISTS $\downarrow$} & \textbf{FID $\downarrow$}& \textbf{KID\footnotemark[1]  $\downarrow$} \\
\midrule
Curr 1 & all & 70.70 & 34.02 & 22.40 & 10.95 & 2.26 \\
Curr 2 & all & 75.47 & \textbf{27.34} & {20.26} & 10.01 & 2.05 \\
Curr 3 & all & 74.50 & 28.45 & 20.99 & 10.20 & 2.09 \\
Curr 4 & all & 73.65 & 29.99 & 21.04 & 10.13 & 1.88 \\
Curr 5 & none & \textbf{75.72} & 27.46 & {20.23} & {9.78} & \textbf{1.76} \\
\textbf{Curr 6} & \textbf{stage 1} & {74.70} & 28.75 & \textbf{20.10} & \textbf{9.73} & 1.82 \\
\bottomrule
\end{tabular}
\label{tab:ablation_curr}
\end{table}

\textbf{Implementation Details}
\label{sec:imple_detail}
All experiments use the AdamW \cite{loshchilov2017decoupled} optimizer with a learning rate of $10^{-5}$.
Due to limited H100 availability, we conduct most experiments on an NVIDIA L40S GPU (45GB). For experiments involving joint training with 
additional components such as ConditionNet, perceptual network, and experiments at $1024\times768$ resolution, we use an NVIDIA H100 GPU (80GB).
Unless noted, we use the dilated masks, masked inputs for IP-Adapter.
We apply data augmentation following IDM-VTON \cite{choi_improving_2025}, including horizontal flip (p = 0.5) and random affine shifting and scaling (p = 0.5), but to the conditioning input only. 
Additionally, color jitter augmentation is applied only in the first stage of training because its unnatural color distribution negatively impacts fine-tuning.
For Leffa, we use the attention module with two heads of dimension 256, comprising 650K parameters, learning rate for attention module is $5\times10^{-3}$ and loss balancing term $\lambda_{\text{Leffa}}=10^{-3}$.
For inference, we use the same inference parameters for all experience: $30$ steps, $CFG=2$, $seed=42$.
Further details in batch size, training time, network's parameters, and inference efficiency are provided in Supp.~Sec.~1
.
Note that for higher resolution, we use the pretrained model of Stage 1 at $512\times384$ to initialize weights.
\subsection{\textbf{Generation Branch}}
All experiments in this branch is conducted within the Stage 1 only.  Qualitative examples are given in 
Supp.~Sec.~4
.

\textbf{Base vs Inpainting Model}
\label{sec:exp_generation_1}
SDXL \emph{base} pipeline consistently outperforms the \emph{inpainting} pipeline across all metrics (Tab.~\ref{tab:architecture_abl}). 
\emph{inpainting} CreationNet slightly misplaces or mis-scales the garment and texture.
These results indicate that VTOFF is primarily a semantic removal task rather than a region completion task. Consequently, the standard SDXL base pipeline achieves a better learning trajectory and local minimum, and we adopt it as the default backbone.

\textbf{SD v1.5 vs SDXL}
\label{sec:exp_generation_2}
Despite comprising substantially more parameters, the higher capacity of SDXL brings significant improvement over SD v1.5 in the same architecture in all evaluation metrics, as shown in Tab.~\ref{tab:architecture_abl}, demonstrating that scaling to SDXL is necessary in the VTOFF setting.
Therefore, we leverage SDXL as the default foundational model for the rest of our experiments.
\subsection{\textbf{Conditioning Branch}}
All experiments in this branch is conducted within the Stage 1 only.  Qualitative examples are given in 
Supp.~Sec.~4
.

\textbf{Truncated vs Full High-level Features}
\label{sec:exp_conditioning_1}
Removing high-level features leads to a severe degradation in all metrics (Tab.~\ref{tab:architecture_abl}). Lack of high-level guidance results in inconsistent identity preservation and incomplete person removal. Interestingly, Try-Off-Anyone \cite{xarchakos_tryoffanyone_2025}, not relying on high-level features, also encounters remaining human body occasionally.
These results confirm that high-level semantic features from the IP-Adapter and text encoder are essential for VTOFF, enhancing contextual understanding.

\textbf{Fit vs Dilated Mask}
\label{sec:exp_conditioning_2}
In general, the dilated mask performs better quantitatively (Tab.~\ref{tab:masking_abl}). 
Fit mask restricts the surrounding context to the garment, probably leading to the lack of generalization on human-garment interaction.

\textbf{Masked vs Unmasked IP Input}
\label{sec:exp_conditioning_3}
In Tab.~\ref{tab:masking_abl}, we witness a marginal difference between both version of masking operation on IP-Adapter input.
\begin{figure}[t]
    \centering
    \includegraphics[width=0.8\linewidth]{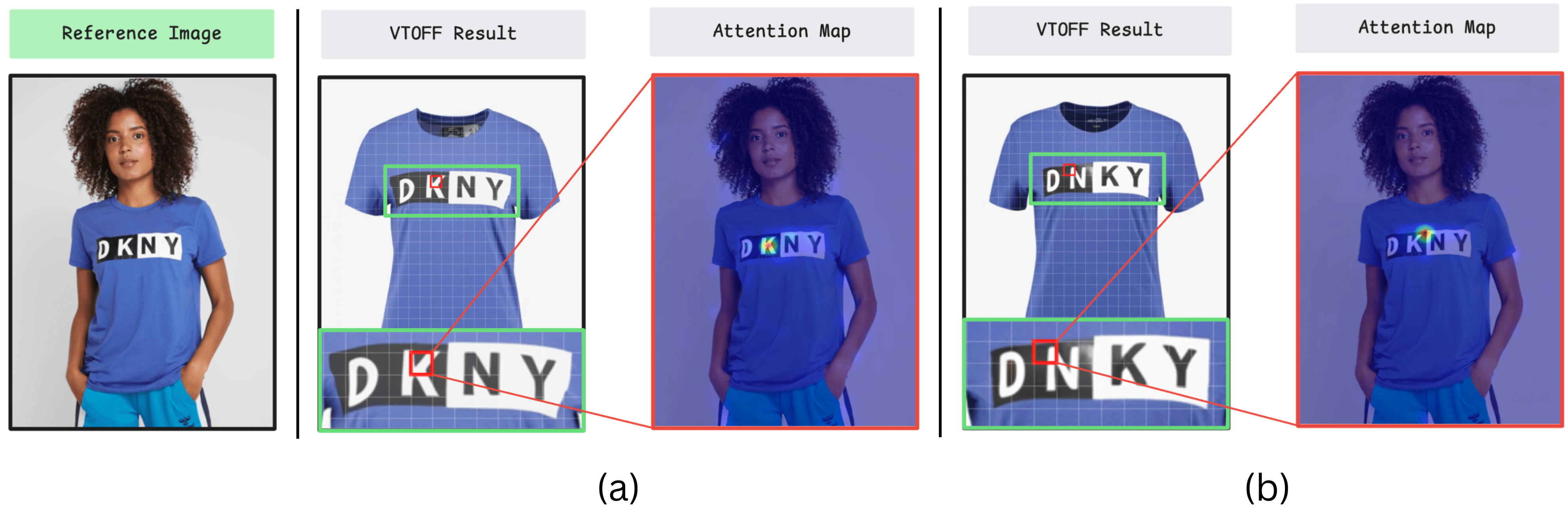}
    \caption{Visualization of Leffa loss impact on guiding attentions toward correct regions and implicitly prevent fine-grained details distortion. \textit{(a)} with Leffa, \textit{(b)} without Leffa
    }
    \label{fig:ablation_leffa}
\end{figure}
\subsection{\textbf{Training Strategies}}
\textbf{Leffa Loss Impact}
\begin{table}[t]
\centering
\caption{Impact of auxiliary losses on VTOFF performance. 
}
\tiny
\begin{tabular}{lccccccc}
\toprule
\textbf{Auxiliary Loss} & \textbf{SSIM $\uparrow$} & \textbf{LPIPS $\downarrow$} & \textbf{DISTS $\downarrow$} & \textbf{FID $\downarrow$}& \textbf{KID\footnotemark[1]  $\downarrow$} \\
\midrule
w/o Leffa & 74.51 & {30.37} & 20.96 & \textbf{10.62} & \textbf{2.22}\\
\textbf{w/ Leffa} & \textbf{75.65} & \textbf{28.15} & \textbf{20.48} & 10.65 & 2.27 \\
\midrule
\textbf{None} & \textbf{75.47} & \textbf{27.34} & \textbf{20.26} & 10.01 & 2.05 \\
ELatent-LPIPS & {74.54} & 28.35 & {20.45} & \textbf{9.96} & \textbf{2.00} \\
LPL & {74.93} & 28.79 & {20.35} & {10.15} & 2.04 \\
\bottomrule
\end{tabular}
\label{tab:ablation_loss}
\end{table}
To validate the effectiveness, we compare the model obtained from Stage 2 with and without Leffa loss.
As illustrated in Tab. \ref{tab:ablation_loss}, the model with Leffa significantly outperforms the model without it across all metrics except a marginal degradation in FID and KID.
Furthermore, when examining the attention map in Fig. \ref{fig:ablation_leffa}, the model with Leffa focuses on the correct regions of interest through spatial regularization, resulting in precise fine-grained details.

\begin{table}[t]
\centering
\caption{Quantitative results comparison with other methods on the VITON-HD (HD) dataset and DressCode (DC) for virtual try-off. $\star$ DISTS is computed at $341\times256$ resolution following TryOffDiff \cite{velioglu_tryoffdiff_2024} evaluation protocol.}
\tiny
\begin{tabular}{lcccccccc}
\toprule
\textbf{Method} & \textbf{Resolution} & \textbf{SSIM $\uparrow$} & \textbf{LPIPS $\downarrow$} & \textbf{DISTS $\downarrow$} & \textbf{FID $\downarrow$}& \textbf{KID\footnotemark[1]  $\downarrow$} \\
\midrule
TryOffDiff \cite{velioglu_tryoffdiff_2024} (HD) & $512\times384$ & \textbf{75.02} & \textbf{28.52} & 22.32 & 24.56 & 9.52 \\
Try-Off-Anyone \cite{xarchakos_tryoffanyone_2025} (HD) & $512\times384$ & 72.35 & 34.08 & 22.11 & 11.57 & 2.01 \\
\textbf{Ours (HD)} & $512\times384$ & {74.70} & 28.75 & \textbf{20.10} & \textbf{9.73} & \textbf{1.82} \\
\midrule
IGR \cite{shen_igr_2024} (HD) & $1024\times768$ & \textbf{78.95} & \textbf{29.46} & 20.45 & 13.14 & 2.97 \\
\textbf{Ours (HD)} & $1024\times768$ & 76.04 & 31.41 & \textbf{19.59} & \textbf{10.69} & \textbf{2.31} \\
\midrule
TryOffDiff \cite{velioglu_tryoffdiff_2024} (DC)$\star$ & $512\times384$ & \textbf{80.8} & \textbf{31.6} & 21.6 & 17.1 & 4.7 \\
\textbf{Ours (DC)$\star$} & $512\times384$ & 75.53 & 32.61 & \textbf{20.85} & \textbf{12.30} & \textbf{2.12} \\
\bottomrule
\end{tabular}
\label{tab:quantitative_res}
\end{table}

\textbf{Curriculum Design}
\label{sec:exp_curr_design}
We design six different training curricula to explore the best training dynamic and stability. 
The varied components consist of the interleaving of CreationNet (with IP-Adapter) and ConditionNet updates, the use of Leffa loss, and the scheduling of color augmentation. The other hyperparameters are maintained as in Sec. \ref{sec:imple_detail}. Using the notations in Sec. \ref{sec:currilearn}, (130) to express number of epochs, and (color) to express the use of color augmentation, our curricula details are as follows:
\begin{itemize}
    \item Curr 1: \emph{joint} (130, color).
    \item Curr 2: \emph{cre\_ip} (130, color) → \emph{cre\_ip} + Leffa (130, color) → \emph{cond} + Leffa (130, color) → \emph{cre\_ip} + Leffa (90, color).
    \item Curr 3: \emph{cre\_ip}(130, color) → \emph{cond} + Leffa (130, color) → \emph{cre\_ip} + Leffa (130, color) → \emph{cond} + Leffa (130, color).
    \item Curr 4: \emph{cre\_ip}(130, color) → \emph{joint} + Leffa (130, color).
    \item Curr 5: \emph{cre\_ip}(130) → \emph{cre\_ip} + Leffa (130) → \emph{cond} + Leffa (130).
    \item Curr 6:  \emph{cre\_ip}(130, color) → \emph{cre\_ip} + Leffa (130) → \emph{cond} + Leffa (130).
\end{itemize}
As shown in Tab.~\ref{tab:ablation_curr}, different curricula produce distinct trade-offs. Curr 1 and Curr 4, which rely heavily on joint training, perform the worst due to training instability from the increase in dynamic variables. Curr 2 provides a strong baseline for LPIPS and DISTS, but its full usage of color augmentation introduces distributional color mismatch, resulting in a fading effect result 
(visualization is given in Supp.~Sec.~4
).
Curr 3 achieves moderate results but does not surpass Curr 2.
Curr 5 and Curr 6 highlight the role of augmentation scheduling. Disabling color augmentation entirely (Curr 5) achieves solid improvements in all metrics, likely due to the elimination of the color fading effect. However, Curr 6 achieves the best balance, yielding the lowest DISTS (the primary metric) and FID while maintaining competitive scores across other metrics.
Crucially, we find that the final fine-tuning for CreationNet (used in Curr 2 and 3) becomes redundant when the color distribution is natural (in Curr 5 and 6), allowing improved efficiency.
This demonstrates that color augmentation in the first stage prevents CreationNet’s overfitting, resulting in a thorough understanding of color and edge, while disabling it in later stages produces natural color and help the model concentrate on preserving fine-grained details.

\textbf{Perceptual Loss Impact}
To evaluate the impact of perceptual supervision, we extend Curr 2 by enabling perceptual losses only in the final stage of training. 
As shown in Tab.~\ref{tab:ablation_loss}, neither ELatent-LPIPS nor LPL improves over the model without perceptual loss. In fact, both objectives slightly harm SSIM, LPIPS, DISTS while providing only marginal gains in FID and KID.
These results suggest that the foundational model capacity is saturated. Adding perceptual supervision introduces extra optimization complexity without clear benefits.

\begin{figure}[t]
    \centering
    \includegraphics[width=\linewidth]{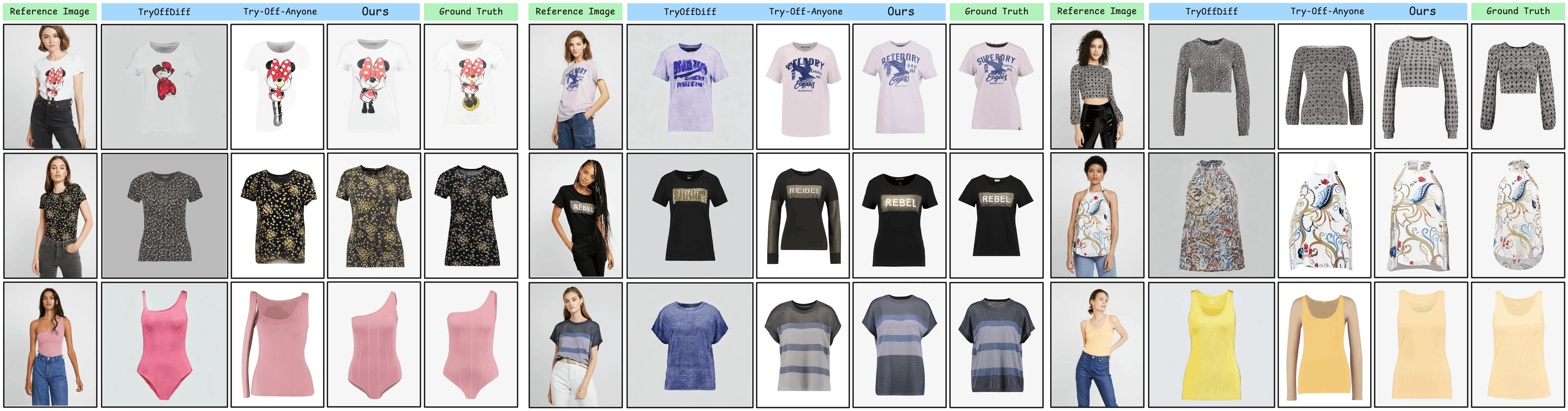}
    \caption{
    Qualitative comparison on VITON-HD dataset between our approach and previous works. Zoom in for better inspection. More examples are given in Supp.~Sec.~6
    .
    }
    \label{fig:qualitative}
\end{figure}

\subsection{\textbf{Comparison with State-of-the-Art}}
\label{sec:compare_sota}
We quantitatively compare our best model (Curr~6, Sec.~\ref{sec:exp_curr_design}) with recent state-of-the-art methods on VITON-HD and DressCode datasets in Tab.~\ref{tab:quantitative_res}. We improve DISTS by 9.5\%, 4.2\% and 9.9\% comparing with Try-Off-Anyone~\cite{xarchakos_tryoffanyone_2025}, IGR~\cite{shen_igr_2024} and TryOffDiff~\cite{velioglu_tryoffdiff_2024} at resolutions $512\times384$, $1024\times768$ and $512\times384$, respectively. Overall, our method achieves the best performance on DISTS, FID, and KID across benchmarks, while SSIM and LPIPS show mixed results. This discrepancy stems from the inherent bias of SSIM and LPIPS toward global structure over local texture; as demonstrated in our supplementary study (Supp.~Sec.~3
), these metrics are sensitive to shape misalignments while under-penalizing the texture distortions that DISTS more accurately captures. For VTOFF, accurate transfer of visible garment textures is more important than correct subtle shape details, such as garment length under occlusion, which commonly occur in VITON-HD and DressCode. Qualitative comparisons (Fig.~\ref{fig:qualitative}) reveal that existing single-UNet approaches fail for different reasons: Try-Off-Anyone, which relies heavily on low-level features, often produces semantically incoherent garment shapes and incomplete human body removal, while TryOffDiff, dominated by high-level features, struggles to transfer fine-grained texture details. Our adapted Dual-UNet explicitly balances high- and low-level feature representations, enabling accurate texture synthesis while preserving coherent garment geometry. This balanced design generalizes robustly to real-world Unsplash images (Fig.~\ref{fig:title_im} third column) under complex backgrounds and occlusions. Additional qualitative results and architectural discussions are provided in the supplementary material.

%% file: sec/5_conclusion.tex
\section{\textbf{Conclusions}}
This work provides 
a rigorous design analysis of Dual-UNet architectures, determining what matters for VTOFF performance. We demonstrated the utility of the Leffa attention regularization module in mitigating fine-grained distortions and introduced a curriculum learning strategy that ensures robust, stable convergence. These key adaptations enabled us to achieve state-of-the-art performance on VITON-HD and DressCode across multiple resolutions, surpassing prior methods on the primary metrics DISTS, and FID, KID. The insights derived from our analysis establish a clear and reproducible framework, advancing current VTOFF performance and guiding the design of future models.

\textbf{Future Works.}
Our evaluation primarily considers single-garment scenarios under commonly studied poses; extending the framework to multi-garment compositions and complex interactions such as multi-layer clothing remains an interesting direction for future work. Moreover, we will explore alternative generative paradigms beyond latent diffusion, such as flow-matching-based models.

%% file: supp/0.tex
\section{Dataset and Training Details}
\label{supp_sec_1}
\begin{figure}[htbp]
    \centering
    \includegraphics[width=0.8\linewidth]{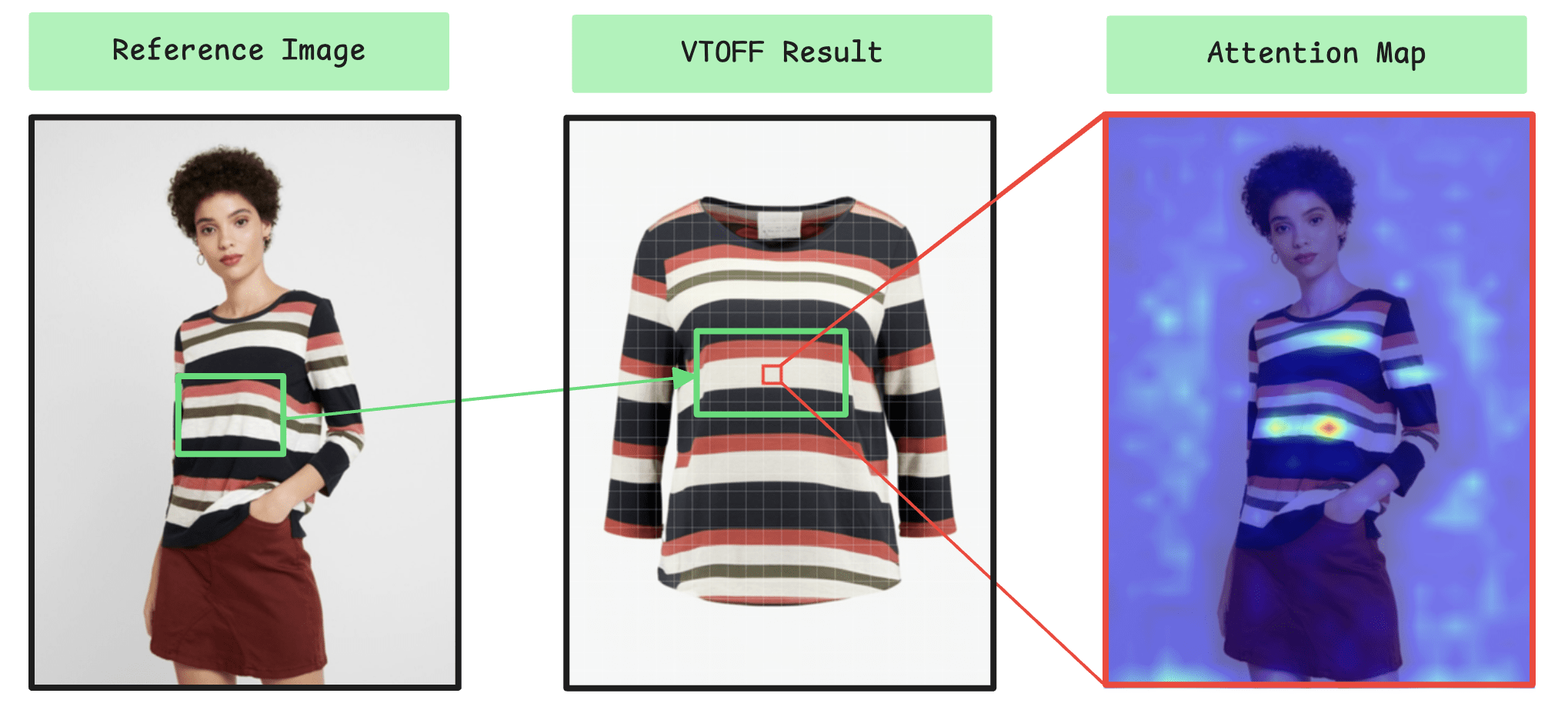}
    \caption{Failure case in VTOFF due to wrong and flat attention similar to VTON \cite{zhou2025learning}. Left: Reference image used in ConditionNet; Center: VTOFF result, which fails to present the olive green stripe; Right: The corresponding attention map of the red region in the result with respect to the reference image.}
    \label{fig:fail_texture_leffa}
\end{figure}
In this section, we provide more details about the dataset, training efficiency and inference hyperparameter:

\textbf{VITON-HD}
VITON-HD dataset \cite{choi2021viton} focuses solely on the upper-body garment type. We follow Velioglu et al. \cite{velioglu_tryoffdiff_2024} to clean duplications to avoid data leakage into the testing set. Finally, the dataset used in training contains 13679 pairs of person and garment images in HD resolution ($1024\times768$) consisting of 11552 training and 1990 testing images.

\textbf{DressCode}
DressCode \cite{morelli2022dresscode} is a multi-garment dataset, which we only utilize the upper-body category in our study. The dataset comprises 48,392 person–garment image pairs in the training split, including 13,563 upper-body, 7,151 lower-body, and 27,678 dress pairs. The test split contains 5,400 pairs, evenly distributed across the three categories (1,800 per category). All images are provided at a resolution of $1024\times768$.

\textbf{Unsplash Real-World} We collect randomly free license photos on Unsplash \cite{unsplash} about fashion models and casual looks in the wild. The set includes 57 images without ground truth. We note that all images are used under the Unsplash License. We then generate mask for upper body following the well-known dilated masking style, which covers the person torso, arms, neck following with a small dilation but keeps the hand, head, and lower body intact.
\begin{table}[htbp]
\centering
\caption{
Trainable parameters, batch size and corresponding GPU usage for each stage in the training curriculum.
}
\scriptsize
\setlength{\tabcolsep}{4pt}
\label{tab:combined_vton_res}
\begin{tabular}{lcccc} 
\toprule
\textbf{Resolution} & \textbf{Stage} & \textbf{Trainable params} & \textbf{Batch Size} & \textbf{GPU} \\
\midrule
\multirow{5}{*}{$512\times384$} & \textit{cre\_ip} & 3.074B & 4 & L40S \\
 & \textit{cre\_ip } + Leffa & 3.075B & 2 & L40S \\
 & \textit{cond} + Leffa & 2.651B & 4 & L40S \\
 & \textit{joint} & 5.642B & 12 & H100 \\
 & \textit{joint} + Leffa & 5.642B & 12 & H100 \\
\midrule 
\multirow{5}{*}{$1024\times768$} & \textit{cre\_ip} & 3.074B & 12 & H100 \\
 & \textit{cre\_ip } + Leffa & 3.075B & 12 & H100 \\
 & \textit{cond} + Leffa & 2.651B & 12 & H100 \\
 & \textit{joint} & 5.642B & 12 & H100 \\
 & \textit{joint} + Leffa & 5.642B & 12 & H100 \\
\bottomrule
\end{tabular}
\label{tab:efficiency}
\end{table}

\textbf{Training Details}
The complete Dual-UNet architecture including IP-Adapter comprises totally 5.6B parameters. However, not all the components are trainable at the same time due to the increasing complexity and ineffectiveness as discussed in the main paper. Moreover, the Leffa attention module adds 650K parameters which are only utilized during training and are disengaged for inference. 
We utilize one NVIDIA L40S GPU for most of ablations and one NVIDIA H100 GPU for heavier experiments. 
Overall, training a stage (130 epochs) in the curriculum requires approximately 50 hours to complete on the L40S GPU and 60 hours at $1024\times768$ resolution on H100 GPU.
Moreover, trainable parameters also vary according to each stage, which Tab.~\ref{tab:efficiency} reports the corresponding amount, batch size used, and GPU.
\begin{table}[htbp]
\centering
\caption{
Inference Resource Consumption using float16 mixed precision on a consumer NVIDIA 4090 24GB GPU with batch size as 1.
}
\scriptsize
\setlength{\tabcolsep}{4pt}
\label{tab:inference_efficiency}
\begin{tabular}{lcc}
\toprule
\textbf{Resolution}& \textbf{Memory} & \textbf{Inference Time (sec/im)} \\
\midrule
$512 \times 384$ & $\sim$16 GB & ~2\\
$1024 \times 768$ & $\sim$23 GB & ~5\\
\bottomrule
\end{tabular}
\end{table}

\textbf{Inference Details}
We use the same inference parameters (30 steps, CFG=2, seed=42) for all experiments. Inference can be executed on the consumer GPU NVIDIA 4090 24GB using float16 mixed precision, with a maximum batch size of 2 at $512 \times 384$ and 1 at $1024 \times 768$ resolution. Tab.~\ref{tab:inference_efficiency} provides the details information about consumed resources when running inference.

%% file: supp/1.tex
\section{Attention Misalignment in VTOFF}
\label{supp_sec_2}
Fig.\ref{fig:fail_texture_leffa} shows a representative failure of VTOFF caused by misaligned and flat attention, a phenomenon previously observed in VTON \cite{zhou2025learning}. In this example, the reference image (left) contains an olive green stripe, but the generated result (center) from the baseline model before we apply Leffa loss fails to reproduce it. The attention map (right) corresponding to the red region in the result indicates weak and spatially diffuse alignment with the reference, hindering accurate transfer of fine-grained garment details. This supports our observation in the main paper that VTOFF suffers from similar attention misalignment issues as VTON, motivating our evaluation of Leffa loss.
\begin{figure}[htbp]
    \centering
    \includegraphics[width=1\linewidth]{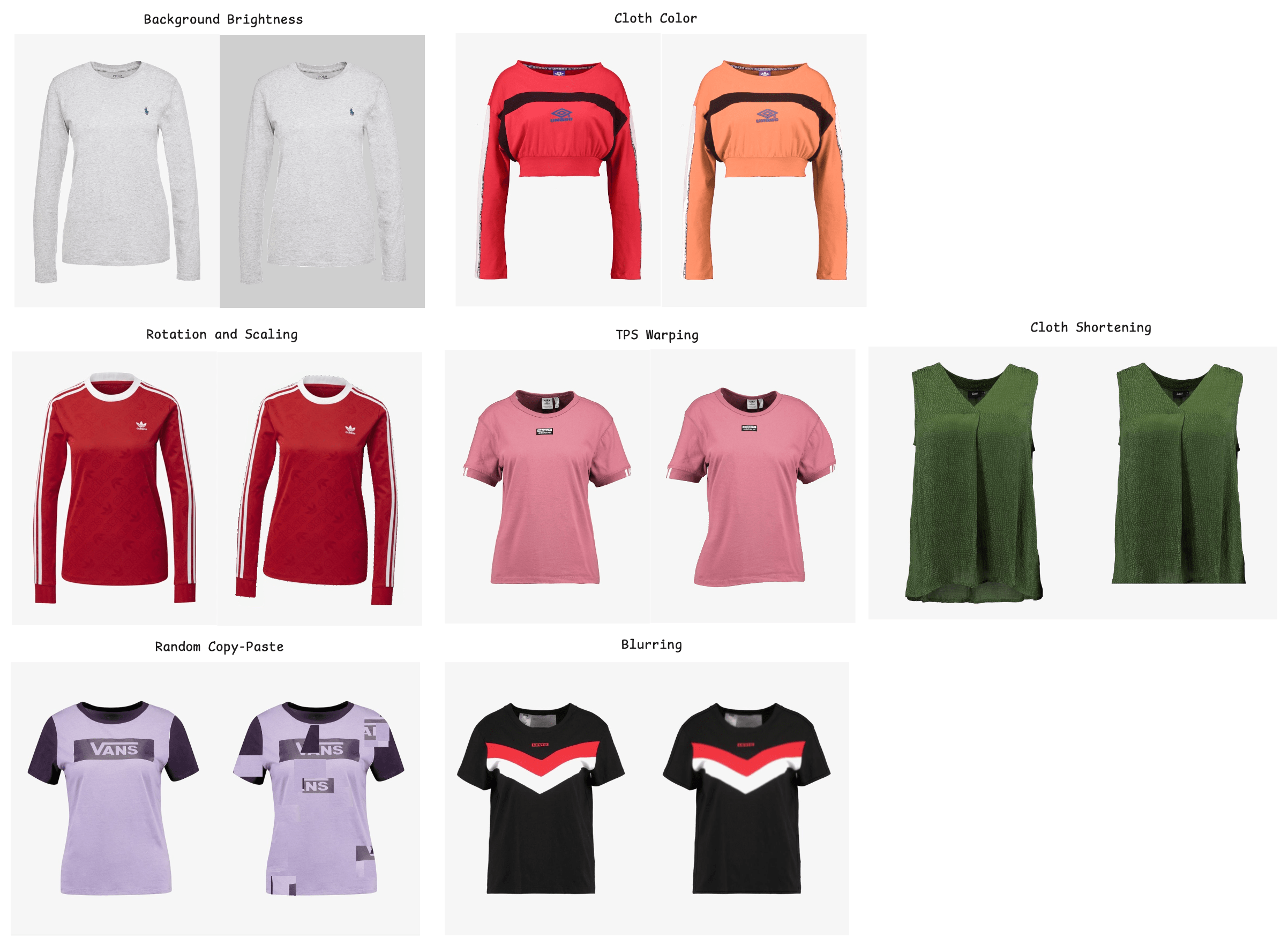}
    \caption{Disentangling transformations applied to the ground truth images of the VITON-HD test set, showing in each pair of images: the left image is ground truth and the right one is the modified image. From top to bottom: color modifications (background, cloth), cloth shape distortions (rotation \& scaling, TPS warping), and mild/drastic texture alterations (random copy-paste, blurring). The reflection on VTOFF metrics of these transformations are provided in Tab.~\ref{tab:metric_exp} and Sec. \ref{sec:exp_metrics}.}
    \label{fig:metric_exp}
\end{figure}

%% file: supp/2.tex
\section{Association between distortions and metrics in VTOFF}
\label{supp_sec_3}
\begin{table*}[htpb]
\centering
\caption{Effect of Background, Color, Shape, and Texture Changes on VTOFF Metrics. Texture changes strongly affect DISTS, FID, and KID, while shape alterations drastically decrease LPIPS and SSIM-based metrics. CLIP-FID is less suitable for VTOFF.}
\begin{tabular}{lccccccc}
\toprule
\textbf{Method} & \textbf{SSIM $\uparrow$} & \textbf{LPIPS $\downarrow$} & \textbf{DISTS $\downarrow$} & \textbf{FID $\downarrow$}& \textbf{CLIP-FID $\downarrow$}  & \textbf{KID\footnotemark[1]  $\downarrow$} \\
\midrule
Background brightness & 99.49 & 1.74 & 3.29 & 1.10 & 1.25 & -0.12 \\
Cloth color & 99.04 & 2.87 & 4.70 & 1.83 & 0.44 & -0.11 \\
Cloth shape & 87.09 & 13.79 & 8.45 & 3.40 & 0.43 & 0.18 \\
Cloth texture (mild) & 96.15 & 8.46 & 9.56 & 7.30 & 0.94 & 2.3 \\
Cloth texture (drastic) & 93.99 & 11.82 & 12.54 & 15.95 & 2.62 & 7.08 \\
\bottomrule
\end{tabular}
\label{tab:metric_exp}
\end{table*}
\footnotetext[1]{KID is multiplied by $10^3$ to bring it to a similar scale as other metrics and is the mean value over 10 evaluation runs.}

To better understand the association of common failures in VTOFF with the metrics, we apply the following set of disentangling transformations (Fig. \ref{fig:metric_exp}) on the ground truth of the test set and observe the metric fluctuation.
\begin{itemize}
    \item Background Brightness: as there is no reference background in which to place the cloth, this is a usual difference between generated images and ground truth. Thus, in the first transformation, we randomly modify the background brightness while keeping the garment intact.
    \item Cloth Color: in contrast to the first experiment, we separate the garment and modify its HSV values. As VTOFF requires transferring the garment, the color should be preserved in the generated image.
    \item Cloth Shape: the prediction of the garment in a new pose can be misshapen. Moreover, in tucking situations, the length of the garment requires hallucination ability. Thus, to simulate these possible failures, we randomly perform one of the following shaping transformations: slight rotation + scaling, shorten the garment at the hem, TPS warping \cite{bookstein_principal_1989}, and re-embed into the original background.
    \item Cloth texture: we randomly perform one of the following transformations: blurring or random copy-paste a portion of the cloth. Additionally, we test 2 levels of distortion magnitude, where the Gaussian kernel's random range for blurring and the number and size of the boxes used to copy-paste are larger in the more severe case.
\end{itemize}

\begin{figure*}
    \centering
    \includegraphics[width=0.7\linewidth]{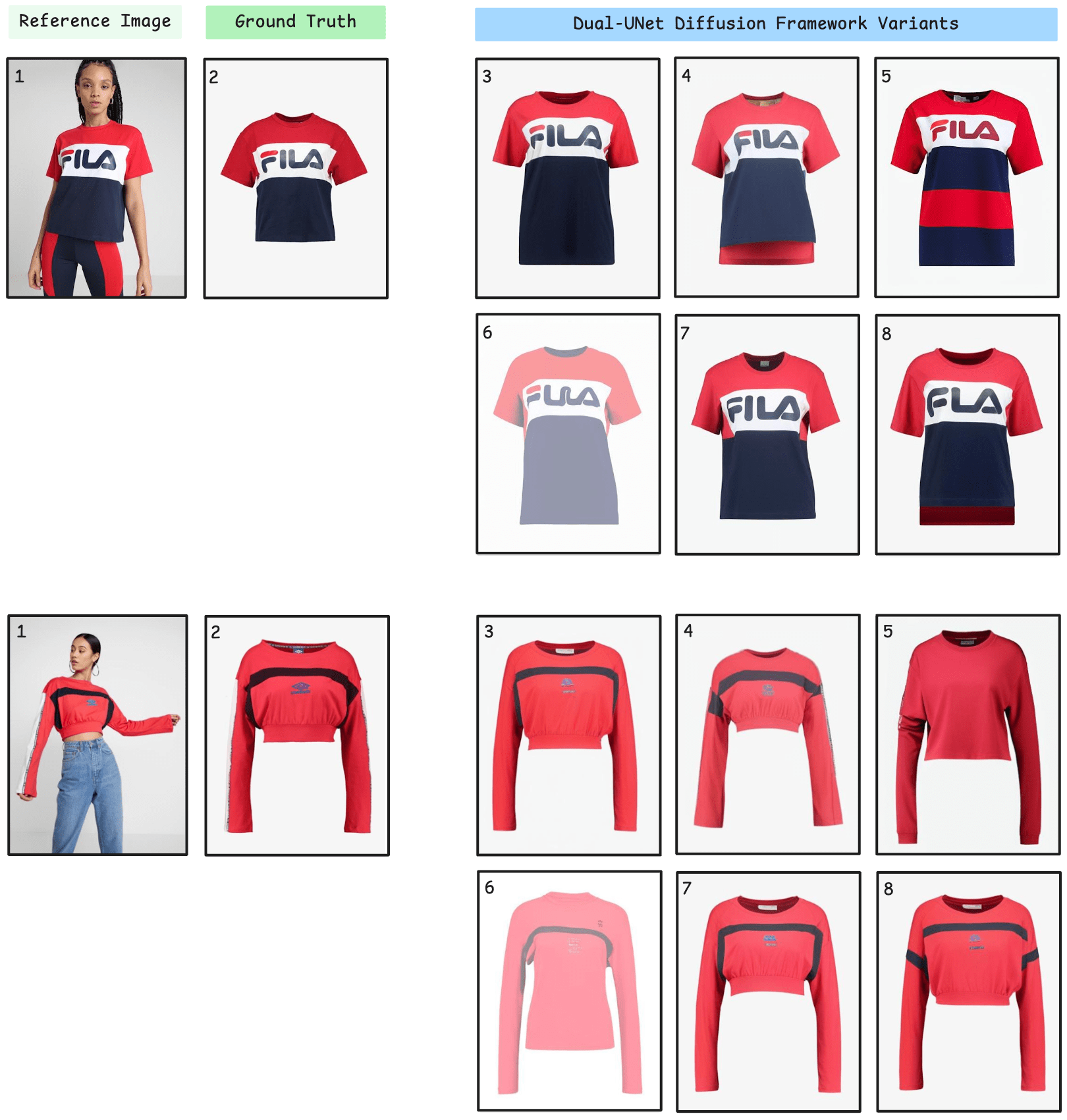}
    \caption{Qualitative comparison of architectural ablations for the proposed Dual-UNet Diffusion Framework on VTOFF. (1) the person wearing the garment, (2) ground-truth canonical garment, (3) the result of the base architecture described in Sec 4.1.3 (main paper), and (4–8) samples of different variants illustrating for Sec 4.2\&4.3 in the main paper: (4)-Inpainting CreationNet in Sec 4.2.1, (5)-SD v1.5 in Sec 4.2.2, (6)-High-level features truncated model in Sec 4.3.1, (7)-Fit mask type in Sec 4.3.2, and (8)-Unmasked input for IP-Adapter in Sec 4.3.3. This illustration clear shows the most suitable architecture for Dual-UNet Framework is the base model as it produces natural high quality image and the identity of cloth is well-preserved.}
    \label{fig:ablation_qualitative}
\end{figure*}

\begin{figure}[htbp]
    \centering
    \includegraphics[width=0.5\linewidth]{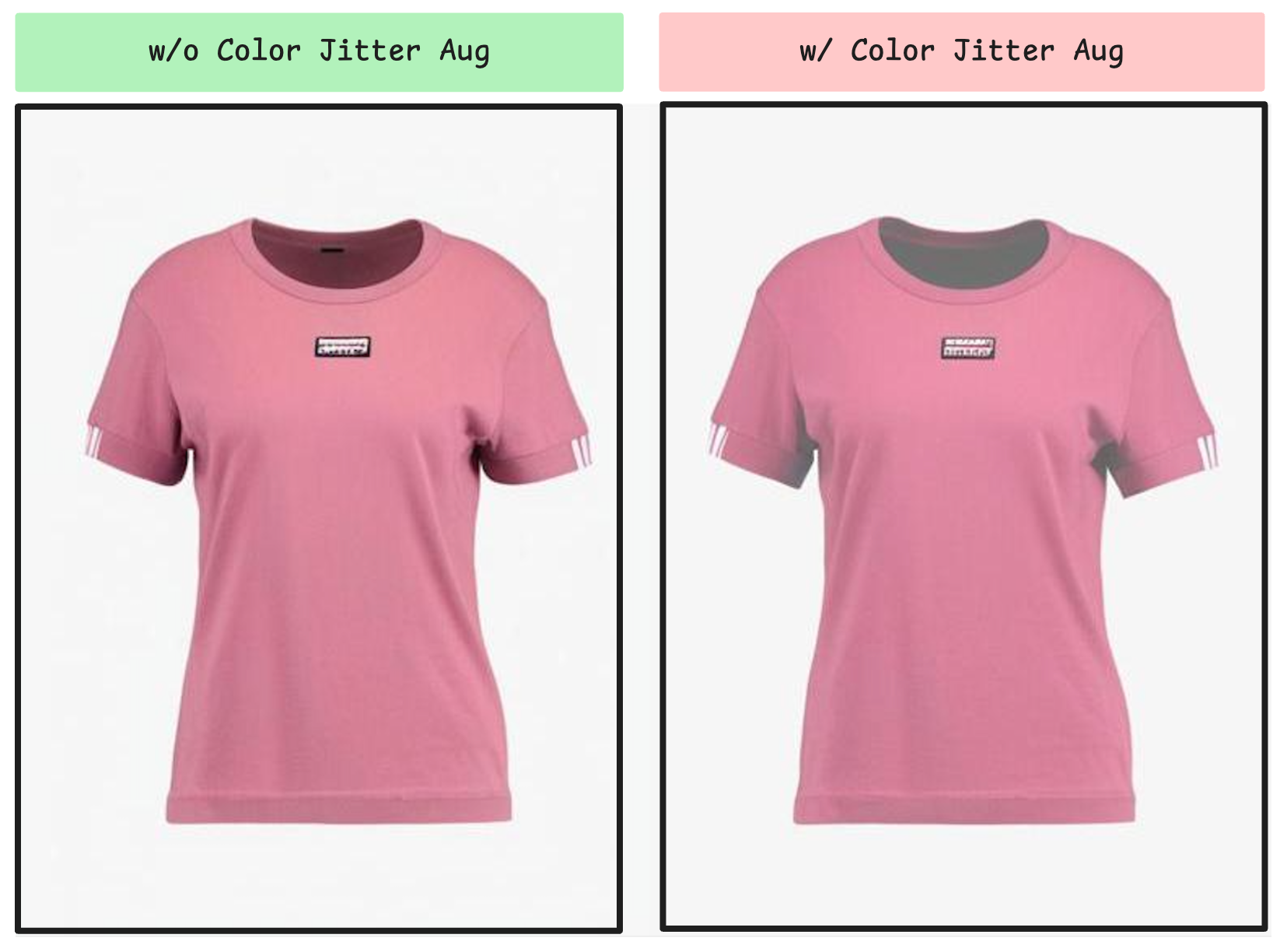}
    \caption{Color fading effect caused by Color Jitter augmentation in later stages of training.}
    \label{fig:color_jitter}
\end{figure}
According to Tab. \ref{tab:metric_exp}, interestingly, a minor misalignment like background brightness results significantly higher CLIP-FID, ranking only behind the drastic change in cloth texture. This indicates that CLIP-FID is not suitable for VTOFF and thus, we exclude it from our experiments.
Second, texture has the most impact on DISTS, FID, KID.
Even a mild distortion in texture, while keeping the shape unchanged, makes DISTS increase dramatically, surpassing than shape modifications with texture preserved.
Besides, LPIPS and SSIM are particularly sensitive to overall shape changes, as drastic changes in cloth texture still have lower LPIPS than changing shape.

%% file: supp/3.tex
\section{Qualitative for architectural ablation study}
\label{supp_sec_4}
To further complement the quantitative results presented in the main paper, we provide qualitative comparisons of different architectural ablations in Fig.~\ref{fig:ablation_qualitative}. Each ablation corresponds to a variant discussed in Secs.~4.2 and 4.3 of the main paper. The figure illustrates how these design choices affect the quality of result in the VTOFF task and confirms the base architecture mentioned in Sec.~4.1.3 (main paper) achieves the most natural and faithful reconstructions.
Furthermore, as illustrated in Fig.~\ref{fig:color_jitter}, the unnatural color distribution in later training stage of curricula prone to distort the color comprehension of model, resulting in a fading effect when generating garment. This insight leads to the decision of removing color jitter augmentation in later phase and yields most natural images.

%% file: supp/4.tex
\section{Random Prompt Augmentation Experiment}
\label{supp_sec_5}
\begin{figure*}
    \centering
    \includegraphics[width=0.75\linewidth]{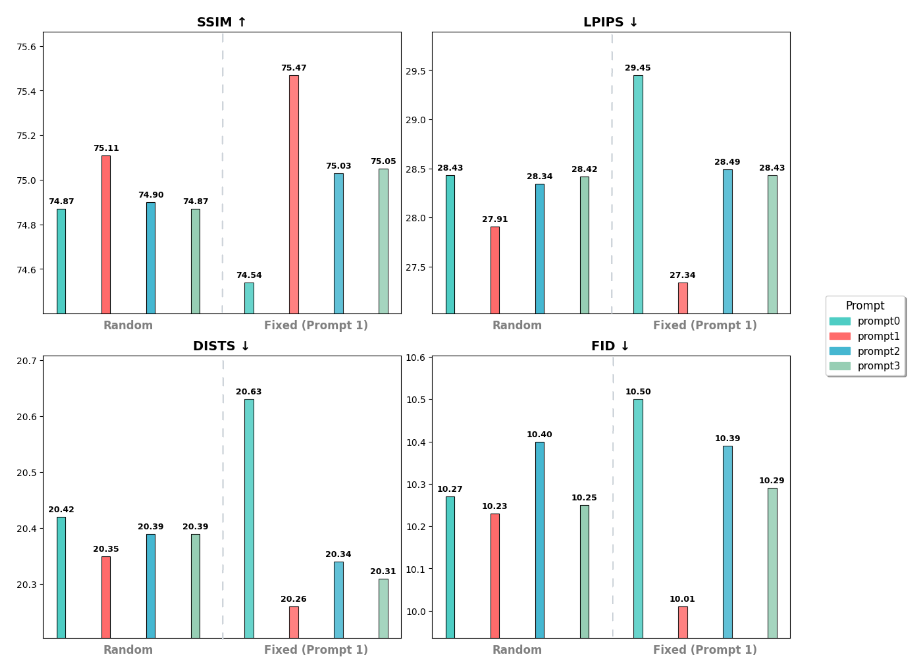}
    \caption{Quantitative comparison of random vs. fixed prompt training across different prompt complexities (ascending from level 0 to 3). Random prompt augmentation yields more consistent performance across prompt levels. Especially, it maintains competitive performance on the shortest prompt - level 0, indicating slightly improved robustness. However, its best performance remains lower than the case where the model is trained with a fixed prompt and evaluated with the corresponding prompt.}
    \label{fig:prompt_result}
\end{figure*}
Textual prompts can vary depending on the user’s writing style. Exposing the model to different levels of prompt complexity during training is expected to improve robustness to textual inputs. To evaluate this effect, we generate three additional levels of prompt complexity using the Qwen 2.5-VL model~\cite{Qwen2.5-VL}, which are randomly sampled during training and compared against the default prompt setting. The four levels are defined as follows:
\begin{itemize}
    \item Level 0: cloth type
    \item Level 1 (also the default prompt): cloth type, neckline, sleeve length
    \item Level 2: cloth type, waist , fit, hem, neckline, sleeve length, cloth length
    \item Level 3: cloth type, waist , fit, hem, neckline, sleeve length, cloth length, color, text if any , textures/patterns
\end{itemize}
To evaluate whether exposure to diverse prompt styles improves robustness in later training stages, we employ a more competitive model. Specifically, we adopt Curriculum 2 (Sec.~4.4.2 in the main paper) and inject the multi-level prompt set during Stage 4. As shown in Fig.~\ref{fig:prompt_result}, this approach slightly enhances robustness, enabling the model to generate consistent results across varying prompt complexities. Nevertheless, it does not reach the performance of a model trained with a fixed prompt and evaluated using the corresponding prompt style during inference.

%% file: supp/5.tex
\section{More Qualitative Results}
\label{supp_sec_6}

\begin{figure*}
    \centering
    \includegraphics[width=0.9\linewidth]{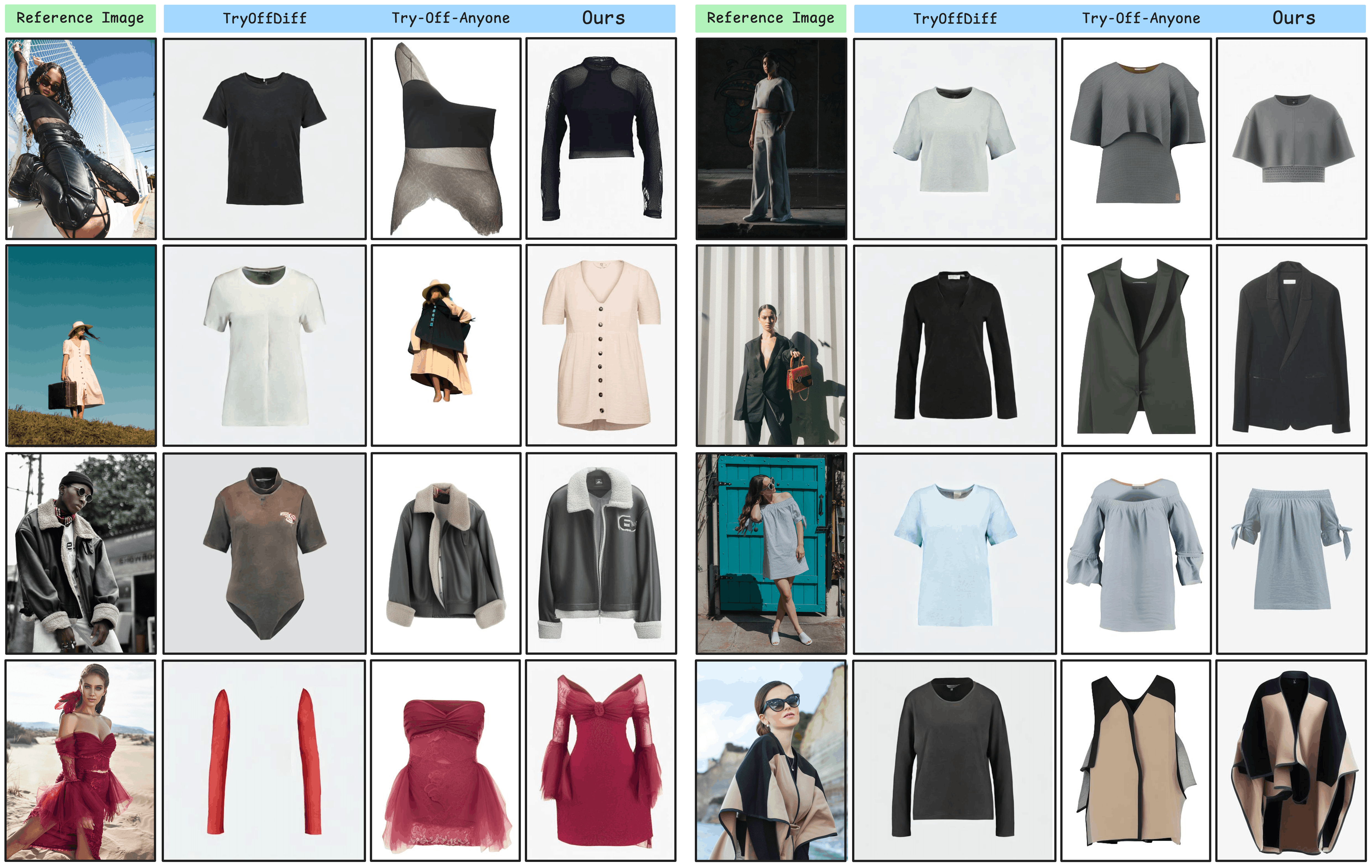}
    \caption{More qualitative comparison with previous works on  at $512\times384$ resolution.
    Qualitative comparison on Unsplash Real-World dataset between our approach and previous works, which are trained on VITON-HD dataset.}
    \label{fig:unsplash}
\end{figure*}

\begin{figure*}
    \centering
    \includegraphics[width=0.9\linewidth]{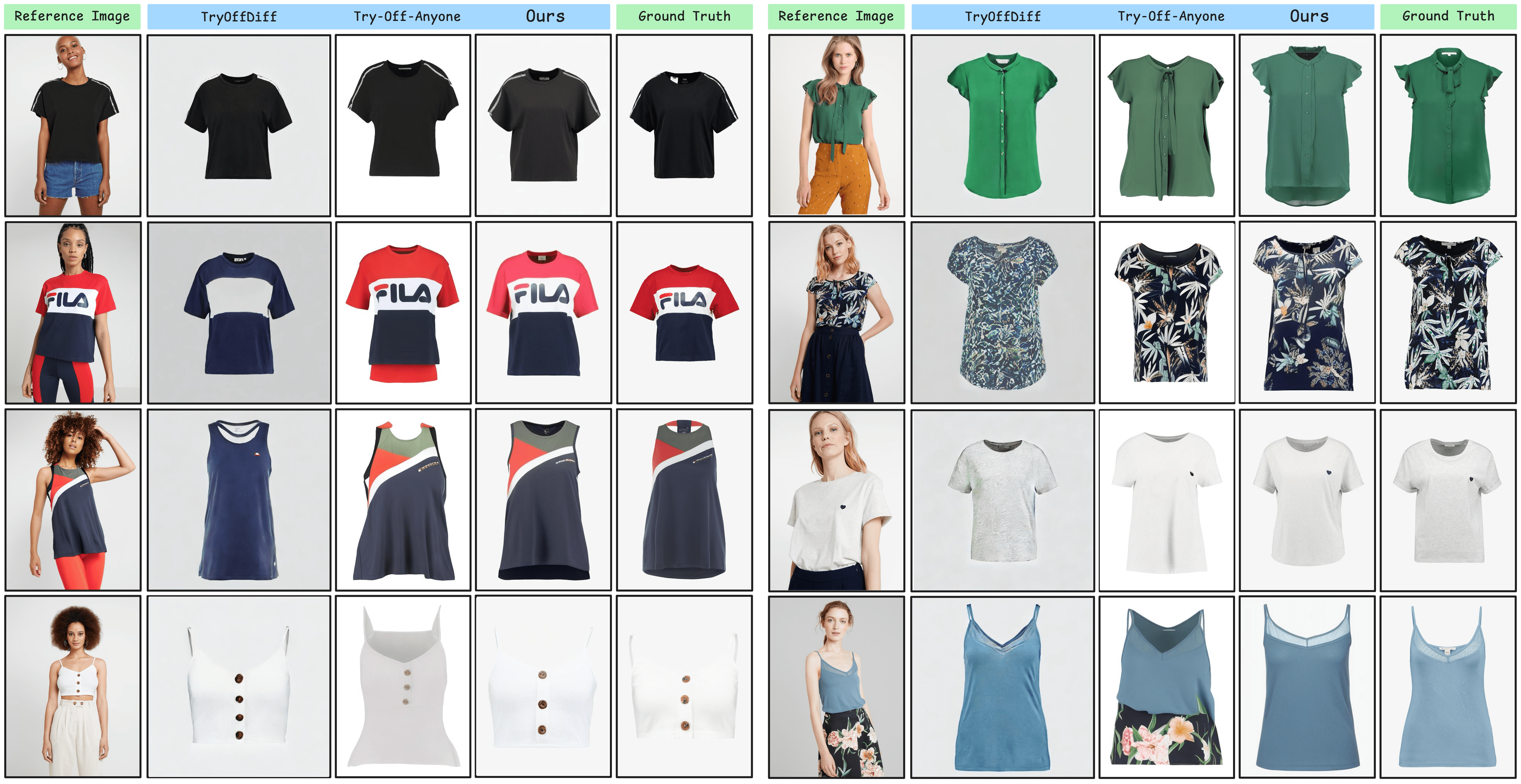}
    \caption{More qualitative comparison with previous works on VITON-HD at $512\times384$ resolution.}
    \label{fig:more_qualitative}
\end{figure*}

\begin{figure*}
    \centering
    \includegraphics[width=0.9\linewidth]{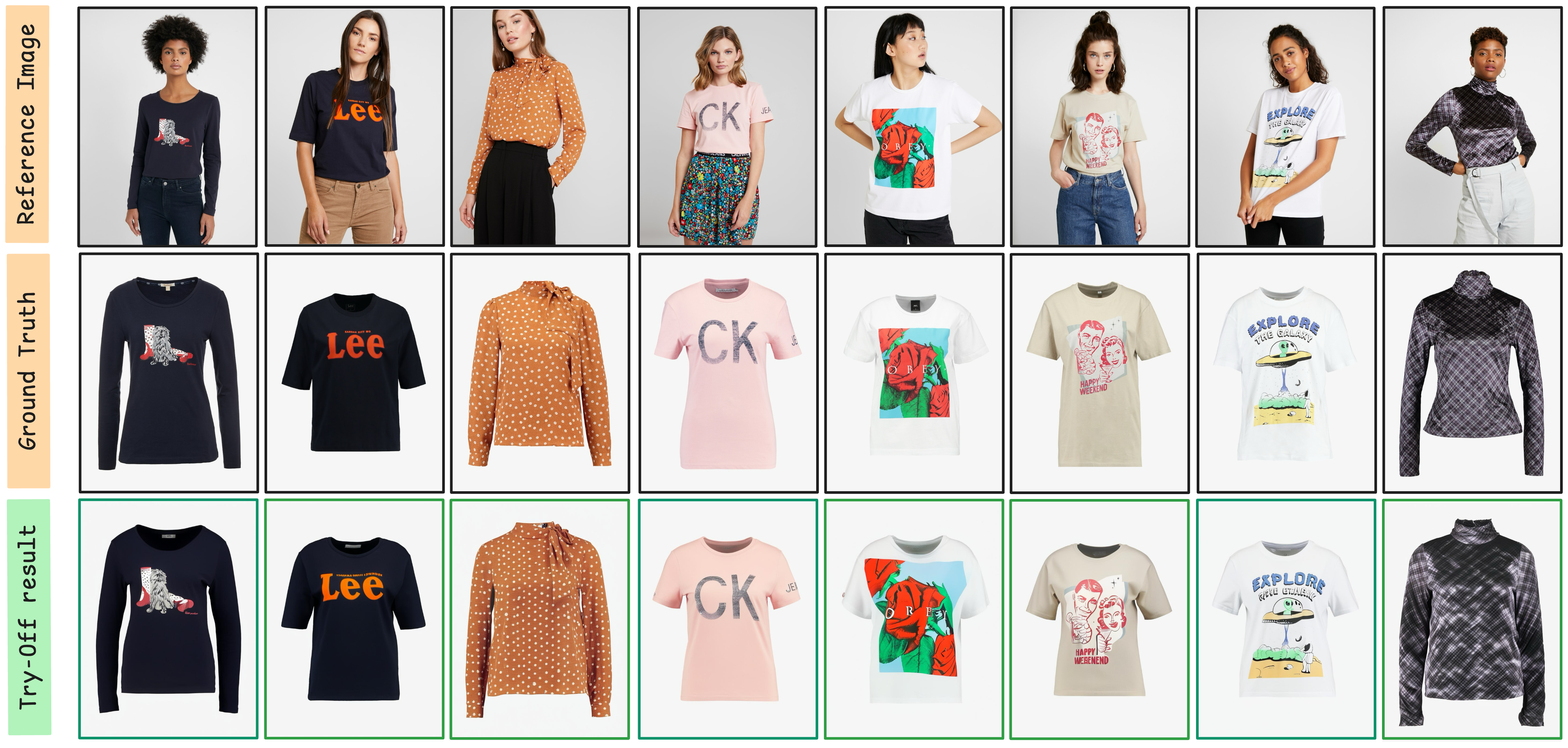}
    \caption{More qualitative examples on VITON-HD at $1024\times768$ resolution produced by our method.}
    \label{fig:more_viton_1024}
\end{figure*}

\begin{figure*}
    \centering
    \includegraphics[width=0.9\linewidth]{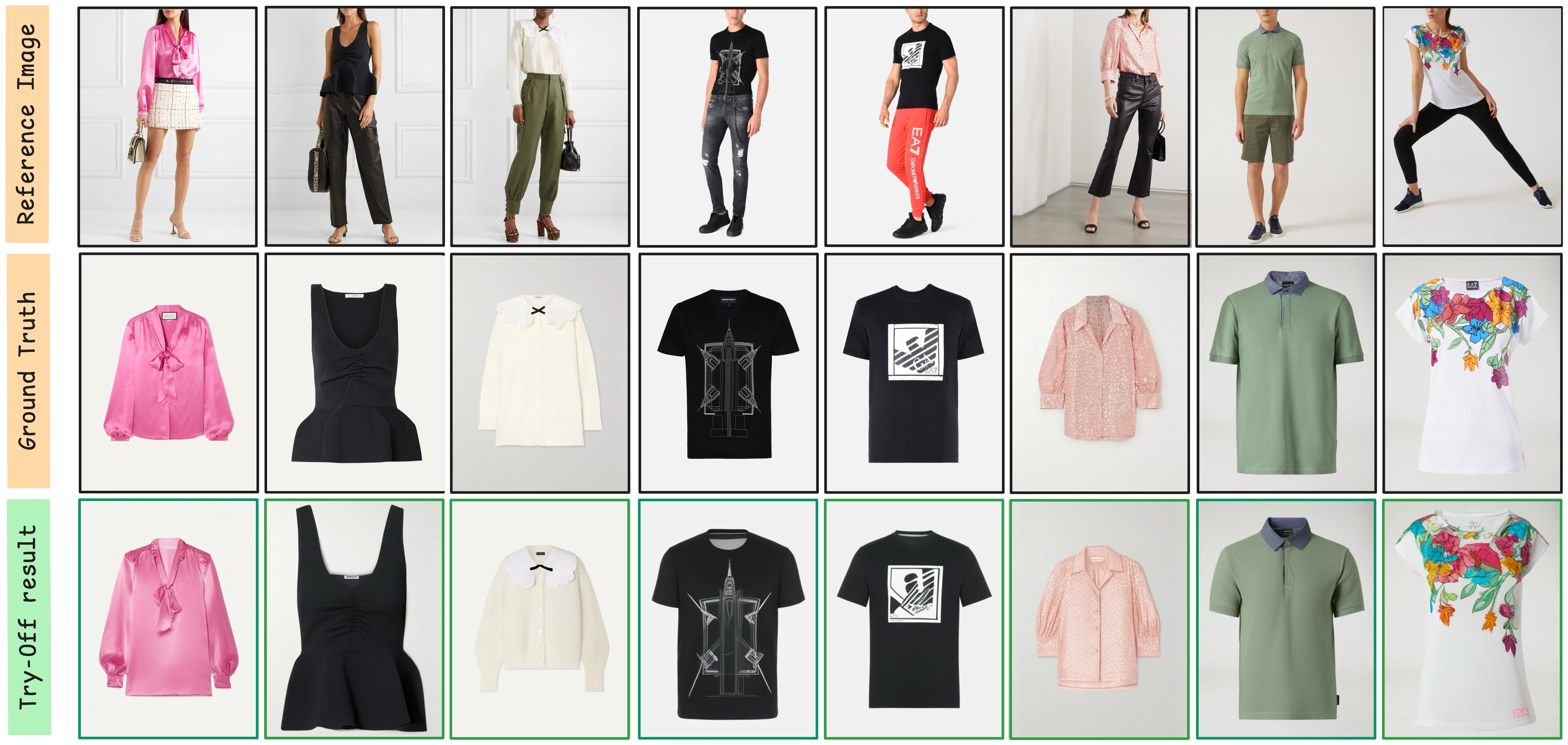}
    \caption{More qualitative examples on DressCode at $512\times384$ resolution produced by our method.}
    \label{fig:more_dc}
\end{figure*}

To complement quantitative evaluations and qualitative results in the main paper, we present additional visualizations. 
Fig. \ref{fig:more_qualitative} provides more results on the VITON-HD dataset at 512×384 resolution, comparing to Try-Off-Diff\cite{velioglu_tryoffdiff_2024} and Try-Off-Anyone\cite{xarchakos_tryoffanyone_2025}. Our approach consistently produces more natural textures and better preserves garment identity.
In Fig. \ref{fig:more_viton_1024}, we provide qualitative results on VITON-HD at 1024×768 resolution. Since IGR\cite{shen_igr_2024} does not publish their source code, only our method is reported here, while their quantitative evaluation is included in the main paper. These examples demonstrate highly garment fidelity and realism, highlighting the ability of our framework to generalize to higher-resolution synthesis.
In Fig. \ref{fig:more_dc}, we provide qualitative results on the DressCode dataset at 512×384 resolution. Try-Off-Diff\cite{velioglu_tryoffdiff_2024} do not offer public implementations for this dataset; hence, only our method is reported, with their quantitative evaluation summarized in the main paper. 
Using model trained on VITON-HD dataset, we further conduct evaluation on Unsplash Real-World samples, presented in Fig.~\ref{fig:unsplash}.
These examples further confirm the robustness of our framework across different datasets, preserving both garment identity and overall visual quality.

\begin{figure}[htbp]
    \centering
    \includegraphics[width=0.7\linewidth]{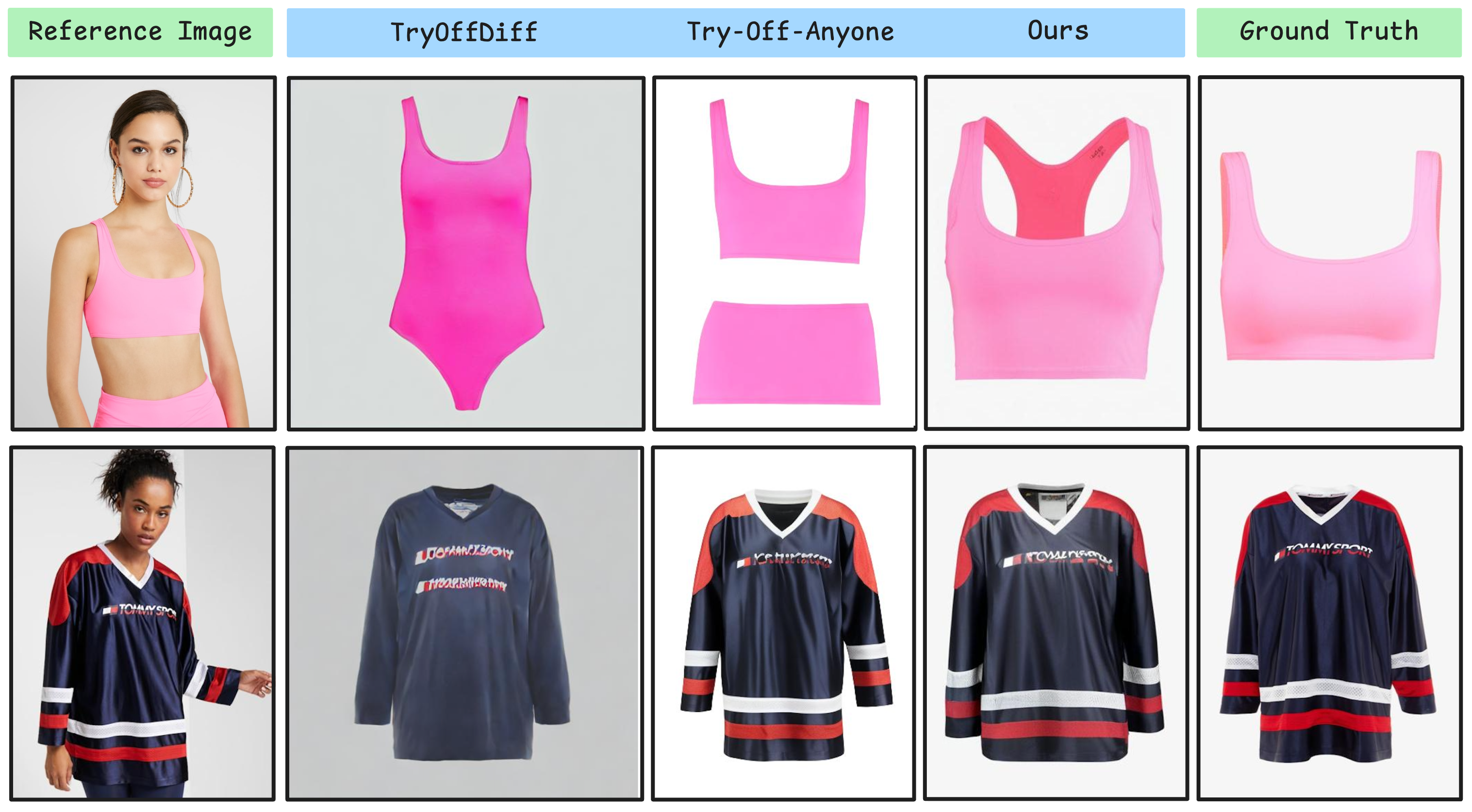}
    \caption{Illustration of Failure Mode. Comparison showing the challenge of maintaining accurate generation for small garment types (top row), and intricate and small text (bottom row).}
    \label{fig:failure}
\end{figure}

%% file: supp/6.tex
\section{Failure Mode Discussion}
\label{supp_sec_7}
Despite achieving state-of-the-art performance, we observe two recurring failure modes that suggest directions for future work. First, the model may occasionally misestimate geometry for small garment types (e.g., bras or gym tops), producing outputs that are slightly longer than the reference. This indicates that incorporating explicit or implicit geometric constraints, such as 3D representation modality, could further improve shape and size accuracy. Second, reconstructing very small and intricate text remains challenging, a known open problem in diffusion-based image generation. Addressing this limitation may benefit from dedicated mechanisms for text control, such as cascaded refinement stages or text-aware auxiliary latents.